\pdfoutput=1
\documentclass{bmvc2k}

\usepackage{amsfonts}
\usepackage[T1]{fontenc}
\usepackage[utf8]{inputenc}

\DeclareMathOperator*{\argmin}{arg\,min}

\usepackage{graphicx}
\usepackage{epstopdf}

\title{Zero-Shot Domain Adaptation via Kernel Regression on the Grassmannian}

\addauthor{Yongxin Yang}{yongxin.yang@qmul.ac.uk}{1}
\addauthor{Timothy Hospedales}{t.hospedales@qmul.ac.uk}{1}

\addinstitution{
RIM Group\\
School of EECS\\
Queen Mary, University of London
}

\runninghead{Yang and Hospedales}{ZSDA via Kernel Regression on the Grassmannian}

\begin{document}

\maketitle

\begin{abstract}
Most visual recognition methods implicitly assume the data distribution remains unchanged from training to testing. However, in practice \emph{domain shift} often exists, where real-world factors such as lighting and sensor type change between train and test, and classifiers do not generalise from source to target domains.
It is impractical to train separate models for all possible situations because collecting and labelling the data is expensive. Domain adaptation algorithms aim to ameliorate domain shift, allowing a model trained on a source to perform well on a different target domain. 
However, even for the setting of unsupervised domain adaptation, where the target domain is unlabelled, collecting data for every possible target domain is still costly. In this paper, we propose a new domain adaptation method that has no need to access either data or labels of the target domain when it can be described by a parametrised vector and there exits several related source domains within the same parametric space. It greatly reduces the burden of data collection and annotation, and our experiments show some promising results.
\end{abstract}

\section{Introduction}

Supervised learning usually assumes that the training and testing data are drawn from the same underlying distribution. This assumption is easily violated in many real-world problems. For example, our classifier might be trained with high-quality images captured by a HD camera in an ideal environment, while the trained model may be applied to images captured in poor lighting condition where the camera is not held still, so the  image is blurred and dark. In this scenario, directly using the pre-trained model  leads to poor performance.

The pervasiveness of this domain shift issue has motivated extensive research in domain adaptation \cite{saenko2010adapting,Gopalan11,gong2012geodesicFlowDA,qui2012domainAdaptDict}. Domain adaptation (DA) aims to undo this distribution shift between the source domain (where the model is trained) and target domain (where the model is applied). DA has two main settings: supervised DA where the target domain has labels but its data volume is very small, and unsupervised DA where the target domain is completely unlabelled. In this paper, we will focus on unsupervised DA -- and eventually a new problem setting of zero-shot domain adaptation -- as they are more practically useful. 

In most DA studies, `domain' is often equivalent to `dataset' for benchmarking convenience.  In the classic Office dataset \cite{saenko2010adapting},  images are split into three discrete domains/datasets based on their capture device. 
However, it is commonly overlooked by the DA community that in many cases \emph{domains are not determined by a single categorical variable, but rather a vector of continuous variables}. For example, consider visual surveillance cameras for person or event recognition: the camera angle $\alpha$ affects the poses of people that it captures, and time of day $\tau$ describes the illumination. In this case domains vary smoothly on a continuous manifold, and $[\alpha,\tau]$ describes a specific domain.  Given that both $\alpha$ and $\tau$ are continuous variables, we have, in fact, an infinite number of domains; and each parameter will have a certain degree of variation that prevents models trained one one domain from working well on another.  Discretizing each factor allows conventional DA to be applied; but this inevitably leads to information loss, and to a large number of domains. We refer to the parameters such as $\alpha$ as domain factors, and the vectors like $[\alpha,\tau]$ as domain descriptors.

With this perspective it is clear that even unsupervised domain adaptation does not scale -- since the number of datasets to collect grows exponentially in the number of domain factors. This motivates our search for a method that can learn from a few domains and then generalise well to an arbitrary parametrised domain \emph{without collecting the data from it}. 
 We refer to this scenario as zero-shot domain adaptation (ZSDA), distinct to supervised and unsupervised domain adaptation. ZSDA is possible if we can predict some pattern (e.g., the subspace that supports the data) of the target domain from the abstract description of the domain descriptor. We formulate this task as a problem of manifold-valued data regression. 
 
 ZSDA aims to enable the attractive use case where a model can consistently and instantaneously perform well on a varying test data distribution, by being automatically `calibrated' on the fly by domain descriptor metadata. This would have important applications in a numerous areas including object \cite{saenko2010adapting}, person \cite{qui2012domainAdaptDict} and audio \cite{yang15} recognition.  To this end, our contributions are three-fold: (i) We propose the novel problem of zero-shot domain adaptation, for  domains parametrised by continuous vectors, (ii) We provide a solution to the proposed problem by designing a novel multivariate regression model for the Grassmannian, (iii) We show promising early results demonstrating the efficacy of  our ZSDA.

\section{Related Work}

\subsection{Zero-Shot Learning}



Zero-Shot Learning (ZSL) has received extensive attention in the computer vision community, such as character \citep{larochelle2008zerodata}, object  \citep{lampert2009learning,fu2014embedding}, and action \citep{liu2011action_attrib} recognition. Instead of building a map (classifier) directly from the image to label space, ZSL studies \citep{larochelle2008zerodata, Palatucci_2009_6459} propose an intermediate representation. This representation may take the form of  `attributes', or more generally, a semantically meaningful descriptor. The motivation is that, assuming the mapping from image to attributes is sufficiently universal, it can be learned from a  large amount of data \cite{socher2013zslCrossModal,yang15}. The mapping can then be used to build recognisers on-the-fly by giving the semantic descriptor for a new object. For example by assigning the attributes  [`black', `white', `stripes'] to the new object  `zebra'.

Inspired by ZSL, we aim to achieve a similar on-the-fly capability for domain adaptation. In the same way that semantic representations such as attributes enable ZSL, we aim to exploit the often freely available domain-metadata to adapt a trained model to any target domain given only its descriptor. The closest work in this area is \cite{yang15}, which first mentioned the problem of zero-shot domain adaptation. However it has the drawback that the domain descriptor is limited to a vector of categorical factors only. We do not have this restriction.

\subsection{Domain Adaptation}

Domain Adaptation (DA) techniques reduce the divergence between source and  target domains such that a model trained on the source performs well on the target. We focus on unsupervised domain adaptation here, as it is more practically valuable. However, it is challenging because we have zero knowledge of the conditional distribution of the target domain $P(Y|X_\mathcal{T})$ thus a discriminative model trained on the source domain $P(Y|X_\mathcal{S})$ can not be leveraged. The remaining option is to exploit the marginal distributions $P(X_\mathcal{T})$ and $P(X_\mathcal{S})$.

There are two main approaches in this area, data and subspace centric. \textbf{Data:} This approach seeks a unified transformation $\phi(\cdot)$ that projects two domains' data into a new space that reduces the discrepancy between the transformed target  $\phi(X_\mathcal{T})$ and source data $\phi(X_\mathcal{S})$. A typical pipeline is to perform PCA \cite{long2013transfer} or sparse coding \cite{long13tsc} on the union of the domains with an additional objective that minimises maximum mean discrepancy (MMD) of the new representations of two domains in a reproducing kernel Hilbert space (RKHS) $\mathcal{H}$, i.e., $\lVert\mathbb{E}[\phi(X_\mathcal{T})]-\mathbb{E}[\phi(X_\mathcal{S})]\rVert_\mathcal{H}^2$, where $\mathbb{E}[\cdot]$ is the expectation operator. \textbf{Subspace:} This approach does not manipulate the data directly, instead it tries to make use of subspaces of two domains. Here, subspace refers to a $D$-by-$K$ matrix of the first $K$ eigenvectors induced by  PCA on the original $D$-dimensional data. We denote $P_\mathcal{S}$ and $P_\mathcal{T}$ as the subspaces of source and target domain learned by two separate PCAs. Subspace Alignment (SA) \cite{fernando2013SA} learns a linear map $M$ for $P_\mathcal{S}$ that minimise the Bregman matrix divergence $||P_\mathcal{S}M-P_\mathcal{T}||_F^2$.  \cite{Gopalan11} samples several intermediate subspaces $P_1, P_2, \dots, P_N$ from $P_\mathcal{S}$ to $P_\mathcal{T}$. That is achieved by thinking of $P_\mathcal{S}$ and $P_\mathcal{T}$ as two points on the Grassmann manifold (Grassmannian) $\mathbb{G}(K,D)$ and finding a geodesic (shortest path on manifold) between them, then the points along the geodesic are  meaningful subspaces. Then all subspaces are concatenated to form a richer linear operator $[P_\mathcal{S},P_1, P_2, \dots, P_N,P_\mathcal{T}]$ that projects two domains into a common space, where the source classifier generalises better to the target domain. A weakness of \cite{Gopalan11} is that the number of intermediate points is a hard-to-determine hyper-parameter. An elegant solution to this, \cite{gong2012geodesicFlowDA} samples all the intermediate points. Although this produces infinitely long feature vectors, excluding conventional linear classifiers, their dot-product is still defined, and thus any  kernelised classifier can be used. A recent study \cite{hoffman2014continuousDA} considered the case where domains are associated with a \emph{single} continuous variable (time) using a sequential PCA and subspace-based DA method. However, it does not extend to a vector domain descriptor, nor to ZSDA.

Reviewing these DA studies, one could easily conclude that target data are compulsory. It initially seems to be impossible to achieve DA without any data in the target domain. However, if we take a deeper look at the subspace approach, the key is the subspace rather than data. If each observed subspace (domain) is associated with an independent vector variable $\mathbf{z}$ (domain descriptor), it is possible to predict a new subspace $P^*$ given its corresponding $\mathbf{z}^*$ via a regression model $\mathbf{z}\rightarrow P$. This is sometimes called manifold-valued data regression problem, where the output space is a manifold (e.g., the Grassmannian) and the input is in Euclidean space. 

\subsection{Manifold-valued Data Regression}

There exists several studies addressing regression in the setting that the independent variable is a point in Euclidean space and the dependent variable is a point in non-flat manifold space such as Riemann and Grassmann manifolds. Based on their methodology, we can group these studies into three categories: (i) Parametric approaches like \cite{Hong2014,journals/ijcv/Fletcher13,conf/cdc/Rentmeesters11,SILVALEITE2013,DBLP:conf/eccv/HinkleMFJ12,Gallivan2003} usually try to find a formulation for the geodesic and then provide a numerical solution for its estimation. (ii) Semi-parametric approach, e.g., \cite{Shi+Styner+Lieberman+Ibrahim+Lin+Zhu2009} uses a link function to map from  Euclidean  to  Riemannian space. (ii) Non-parametric approaches such as  \cite{Davis+Fletcher+Bullitt+Joshi2007, Davis+Lazebnik2008} adapt kernel regression to the manifold case by observing that they are all essentially about searching for a point for which the sum of its (reweighed) distances with all training points is minimised.

Note that for most parametric solutions, the independent variable is  assumed to be a scalar (univariate regression). This is because: (i) In  applications where these methods are popular, e.g., medical imaging, one usually wants to find a pattern against a single factor (e.g., age) and (ii) it is technically challenging to extend the method to multivariate case \cite{DBLP:conf/cvpr/KimBACCJDS14}, because the prediction no longer corresponds to a single geodesic curve, which makes the gradient derivation problematic.

We therefore aim to find a solution based on a non-parametric method since the kernel function usually does not make assumptions on whether the input is a scalar or a vector.     

\section{Methodology}

\subsection{Kernel Regression on Grassmannian}\label{sec:KRG}
\label{krg}
Our goal is to build a regression model that takes an $M$-dimensional vector of independent variables as the input and predicts a point on Grassmannian (represented by a matrix with orthonormal columns).  The output constraint means it can not be treated as conventional Euclidean regression. Therefore, we design a regression model for the Grassmannian. 

\subsubsection{Kernel Regresison Review}

We first review kernel regression. Assume we are given a set of (data, label) pairs,

\begin{equation}
\{(\mathbf{z}_1,P_1), (\mathbf{z}_2,P_2), \dots, (\mathbf{z}_N,P_N)\}
\end{equation}

\noindent where $\mathbf{z} \in \mathbb{R}^M$ and $P \in \mathbb{R}^1$; and a kernel function $k(\mathbf{z}_1,\mathbf{z}_2)$ that measures the similarity between $\mathbf{z}_1$ and $\mathbf{z}_2$. The kernel regression prediction of a test point $\mathbf{z}$ is then estimated by,

\begin{equation}
\label{eqkr}
P = \frac{\sum_{i=1}^{N} k(\mathbf{z},\mathbf{z}_i)P_i}{\sum_{i=1}^{N} k(\mathbf{z},\mathbf{z}_i) }.
\end{equation}

\subsubsection{From Euclidean to Grassmannian regression}

When $P \in \mathcal{M}$ where $\mathcal{M}$ is a non-flat manifold and $P$ is no longer a scalar, Eq.~\ref{eqkr} can be invalid. For example, suppose $\mathcal{M}$ is a Grassmannian $\mathbb{G}(K,D)$ so its members are now matrices $P\in\mathbb{R}^{D\times K}$ with constrains that $P^TP = I_K$. Eq.~\ref{eqkr} can be applied and matrices $P_i$ added, but this is meaningless because adding two points on the Grassmann manifold does not necessarily give another point on Grassmann manifold.

Inspired by \cite{Davis+Fletcher+Bullitt+Joshi2007}, we propose to think of kernel regression as the solution of the following optimisation problem:

\begin{equation}
\argmin_{P\in\mathbb{R}^1} \sum_{i=1}^{N} w_i(P-P_i)^2
\end{equation}

\noindent where $w_i=\frac{k(\mathbf{z},\mathbf{z}_i)}{\sum_{i=1}^{N} k(\mathbf{z},\mathbf{z}_i) }$. More generally, we have

\begin{equation}
\argmin_{P} \sum_{i=1}^{N} w_i \operatorname{d}^2(P,P_i)\label{eq:generalOpt}
\end{equation}

\noindent where $\operatorname{d}^2(\cdot,\cdot)$ is a metric (distance function). $P$ is the Fréchet mean if the minimizer is unique (or Karcher mean when it is a local minimum). Note that the Fréchet mean is defined in general metric space, thus it provides a way to work with manifold-valued data as long as we can find a well defined distance function for the points on the manifold.

\paragraph{Grassmann Manifold Background}
We first review some basic concepts about the Grassmannian. Many distances on the Grassmannian are defined based on a key concept called `principal angle', which can be calculated by SVD. E.g., for two points $P_1$ and $P_2$ on $\mathbb{G}(K,D)$,  

\begin{equation}
P_1^TP_2 = USV^T
\end{equation}

\noindent where $S = \operatorname{diag}(\cos(\theta_1), \cos(\theta_2), \dots, \cos(\theta_K))$. The angle $\theta_k = \cos^{-1}(S_{k,k})$ is the $k$th principal angle.
Table~\ref{tab:dg} lists some frequently used distance functions on the Grassmannian. The details on deriving these functions and their comparison are beyond the scope of this paper. Readers who are interested in this topic should see \cite{DBLP:conf/icml/HamL08} and \cite{2014arXiv1407.0900Y} for a good reference.

\begin{table}[h]
\caption{Distances $\operatorname{d}^2(P_1,P_2)$ on $\mathbb{G}(K,D)$ in terms of principal angles and orthonormal bases}\label{tab:dg}
\renewcommand{\arraystretch}{1.75}
\centering
\begin{tabular}{lll}
\hline & Principal angles & Orthonormal bases\\
 \hline
Binet--Cauchy distance & $1 - \prod\nolimits_{k=1}^K\cos^2\theta_k$ & $1 - (\det( P_1^\mathsf{T} P_2))^2$\\
Chordal distance & $\sum\nolimits_{k=1}^K\sin^2 \theta_k$ & $\frac{1}{2}\lVert P_1P_1^\mathsf{T} - P_2P_2^\mathsf{T} \rVert_F^2$\\
Martin distance & $\log \prod\nolimits_{k=1}^K (\cos^2 \theta_k)^{-1}$ & $-2 \log(\det(P_1^\mathsf{T} P_2))$\\
Procrustes distance & $ 4\sum\nolimits_{k=1}^K\sin^2\frac{\theta_k}{2}$ & $\lVert P_1U - P_2V \rVert_F^2$\\
\hline
\end{tabular}
\end{table}

\paragraph{Manifold-valued data regression with vector input}
For our manifold-valued data regression task, we choose Binet--Cauchy (BC) distance, because of its favourable sensitivity properties \cite{DBLP:conf/icml/HamL08}, and because it is amenable to deriving gradients. 
Substituting the BC distance into Eq.~\ref{eq:generalOpt}, we obtain the following objective function to optimise:

\begin{equation}
\argmin_{P\in\mathbb{R}^{D\times K}} 1-\sum_{i=1}^{N} w_i (\det(P^TP_i))^2,\label{eq:bcOpt}
\end{equation}

\noindent which is subject to constraint $P^T P = I_K$. The gradient with respect to $P$ is,

\begin{equation}
\nabla_P = \sum_{i=1}^{N} - w_i(\det(P^TP_i))^2 P_i (P^TP_i)^{-1}.
\end{equation}

\noindent Vanilla gradient descent is not applicable because of the orthogonality constraints. It is a non-trivial optimisation problem as the constraints lead to non-convexity. A simple solution is to do gradient descent as usual and re-orthogonalise the matrix after each step, but it is numerically very expensive. There are some studies on this topic, such as \cite{1998quantph6030T}, \cite{MarkD04}, and \cite{2014arXiv1407.5965S}. We adopt the solution from \cite{OptM-Wen-Yin-2010}. It applies a Crank-Nicolson-like update scheme that preserves the constraints.

\noindent Given a feasible point  $P$ and the gradient $G=\nabla_P$, a skew-symmetric matrix $A$ is defined as,

\begin{equation}
A := GP^T - PG^T.
\end{equation}

\noindent The new trial point is determined by the Crank-Nicolson-like scheme,

\begin{equation}
P^{\text{Update}}(\eta) = P - \frac{\eta}{2}A(P+AP^{\text{Update}}(\eta))
\end{equation}

\noindent where $\eta$ is the learning rate, and $P^{\text{Update}}(\eta)$ is given by the closed form,

\begin{equation}
P^{\text{Update}}(\eta) = QP
\quad\text{where}\quad
Q = (I+\frac{\eta}{2}A)^{-1}(I-\frac{\eta}{2}A).
\end{equation}

\noindent The details of deriving equations (8) -- (10) can be found in \cite{OptM-Wen-Yin-2010}. 

\subsection{Zero-Shot Domain Adaptation}

Using the methodology developed in Sec.~\ref{krg}, our ultimate goal of zero-shot domain adaptation becomes possible. 
Assuming that we have $N$ observed domains given by

\begin{equation}
\{(X_1, [\mathbf{y}_1], \mathbf{z}_1),(X_2, [\mathbf{y}_2], \mathbf{z}_2),\dots,(X_N, [\mathbf{y}_N], \mathbf{z}_N)\},
\end{equation}

\noindent where $X_i$ is the feature matrix, from which we can learn a subspace $P_i$ by PCA. $\mathbf{z}_i$ and $\mathbf{y}_i$ are the domain descriptor and label vector respectively for domain $i$. $[\mathbf{y}_i]$ indicates that we do not assume all observed domains have been labelled. 

For an unseen domain with descriptor $\mathbf{z}^*$, we can predict its subspace $P^*$ based on the proposed method (Eq.~\ref{eq:bcOpt}) and the training data

\begin{equation}
\{(\mathbf{z}_1, P_1),(\mathbf{z}_2, P_2),\dots,(\mathbf{z}_N, P_N)\}.
\end{equation}

\noindent Once $P_*$ is obtained, any subspace-based DA method (e.g., \cite{fernando2013SA, Gopalan11, gong2012geodesicFlowDA}) can be applied to align the unseen (target) domain to any labelled source domain where a classifier was trained.

\section{Experiments}

\noindent\textbf{Dataset:}\quad  To test our algorithm, we need a dataset which exhibits a range of continuously parametrised  domains (Sec.~\ref{sec:KRG}). However, most existing DA and more general vision datasets are grouped into discrete domains/datasets. We therefore alter an existing dataset for our purposes. We use the Office Dataset \cite{saenko2010adapting}, which collects the images of office supplies from three sources: \emph{Amazon}, \emph{webcam} and \emph{DSLR}. It is a classic dataset to evaluate  domain adaptation algorithms. The typical experimental design for this dataset is to evaluate recognition performance when a model is trained on one domain (e.g., Amazon) and tested on another (e.g., webcam). To test our algorithm, we create a new dataset based on Office.

\begin{figure}
\centering
\includegraphics[width=0.7\linewidth]{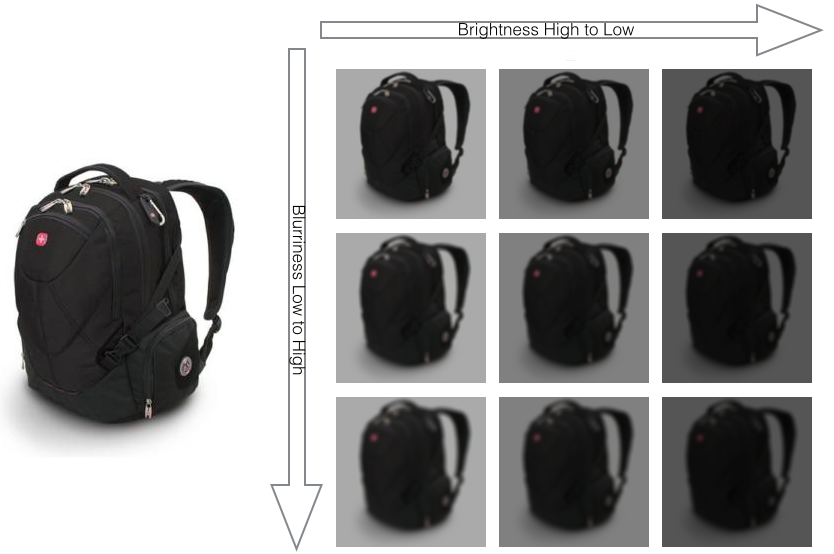}
\caption{Example images in category Backpack. (Left) Original. (Right) Nine simulated domains}\
\label{demoimages}
\end{figure}

\noindent\textbf{Settings:}\quad We use all the Amazon images, which contain 31 categories and with an average of 90 images each. Then, we simulate a range of continuously parametrised domains by degrading each image by two means: lowering the resolution and reducing the brightness. Specifically, we apply a Gaussian filter for simulating low resolution and divide every pixel value by a factor for simulating poor lighting. The size of Gaussian filter and the darkening factor provide two factors of the domain descriptor. In this experiment, we generate nine distinct domains by three levels of degradation for each parameter as shown in Table~\ref{ninedomain}. An example of each domain along with the original image is shown in Fig.~\ref{demoimages}.  For each domain, we split the training and testing set equally, and the assignment of  train versus test set is consistent for all domains. This guarantees that when an image appears in training set, its other versions of degradation will not appear in the testing set. 

\begin{table}[h]
\caption{Nine domains generated by the degradation of Office/Amazon.}
\label{ninedomain}
\centering
\begin{tabular}{lccccccccc}
\hline Domain Index & 1 & 2 & 3 & 4 & 5 & 6 & 7 & 8 & 9 \\ 
Gaussian Filter Size & 5 & 5 & 5 & 10 & 10 & 10 & 15 & 15 & 15 \\ 
Brightness Factor & 1.5 & 2 & 3 & 1.5 & 2 & 3 & 1.5 & 2 & 3 \\ 
\hline 
\end{tabular} 
\end{table}

\noindent\textbf{Features:}\quad We use the state-of-the-art Convolutional Neural Network (CNN) model VGG-full \cite{Chatfield14} as the feature extractor. The image is first preprocessed: rescaled into 224$\times$224 and mean subtracted. Then it is fed into the CNN, where the value in the penultimate layer (4096 neurons) is used as the feature vector for further experiments.

\subsection{Demonstrating the Domain Shift Challenge}
In the first experiment, we demonstrate how  domain shift effects recognition performance, and to what degree domain adaptation methods can alleviate it. A linear SVM classifier is trained on the original domain's training data and we evaluate the performance of the classifier measured by accuracy ($\frac{\textbf{Num. of Correctly Classified Examples}}{\textbf{Num. of All Examples}}$) in the nine degraded domains' test data. Then we apply a popular domain adaptation algorithm -- Geodesic Flow Kernel (GFK) \cite{gong2012geodesicFlowDA} that aims to manipulate the target subspace so that  domain shift is reduced. 

\begin{figure}[t]
\centering
\includegraphics[width=1.01\linewidth]{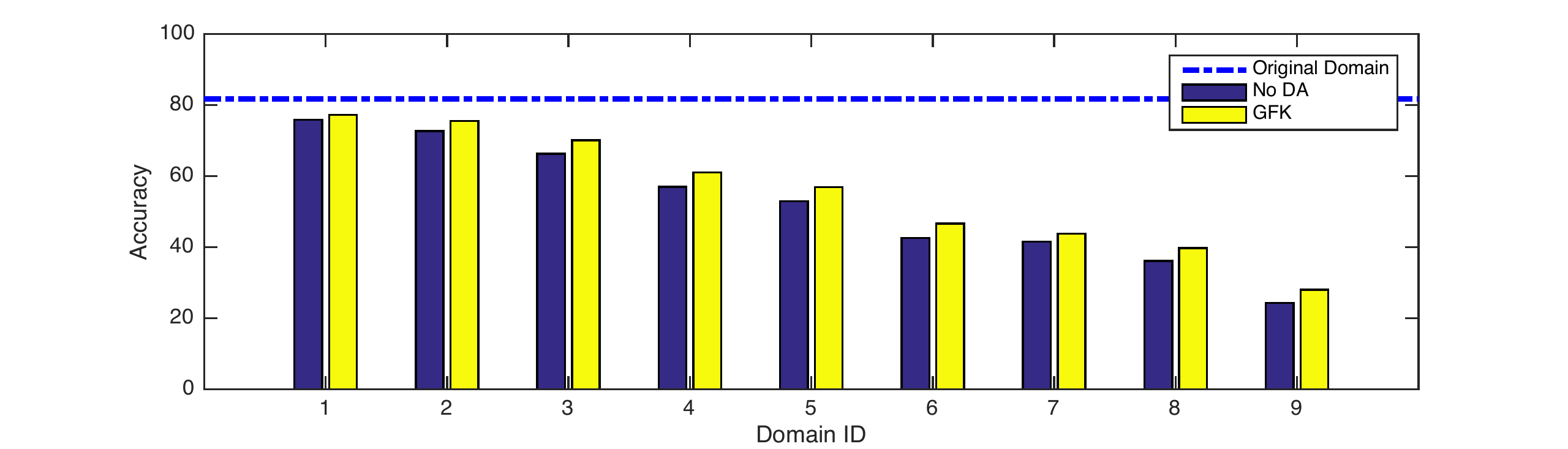}
\caption{Recognition rate in Office/Amazon. The domain shift problem is significant despite the use of state of the art CNN features. }\label{res1}
\end{figure}

As we can see in Fig.~\ref{res1}, the CNN feature is very discriminative. The accuracy in the original domain's test data is 81.68\% (blue horizontal line in Fig.~\ref{res1}). However, the performance inevitably drops when the quality of images gets lower. For the last domain, domain\_9, the accuracy has declined to 24.29\%. Nevertheless, unsupervised domain adaptation by GFK does  reduce the performance drop. It improves the recognition rate by 7.42\% on average.

From these results we conclude that: (i) state of the art features do not eliminate the domain-shift problem (contrary to some claims \cite{Donahue_ICML2014}), and (ii) domain adaptation methods still play an important role in the era of deep learning.

\subsection{Zero-Shot Domain Adaptation}
In the second experiment, we validate the proposed algorithm's ability to undo domain shift given only a target domain descriptor.

\vspace{0.1cm}\noindent\textbf{Target domain performance for different sources of labels:}\quad
The fifth domain, domain\_5, is chosen to be the target domain, and the evaluation is run on its testing data. Unlike traditional domain adaptation approaches \cite{gong2012geodesicFlowDA}, \emph{none of the data (either feature or label) in target domain is given}. All the known information about the target domain is the  descriptor of domain\_5, i.e., $\mathbf{z}_5=[10, 2]$. Besides this, we assume all the training data from the other domains as well as their domain descriptors are given, from which we can learn 8 subspaces via PCA (the reduced dimension is $K=512$). Then the subspace of domain\_5 is predicted by the proposed algorithm so that the subspace based domain adaptation \cite{gong2012geodesicFlowDA} can be applied. Note that we rescale each factor in $\mathbf{z}$ to $[0,1]$ to reduce the effect of different ranges of factors, and RBF kernel with $\sigma=0.1$ is chosen to be $k(\cdot,\cdot)$. 

\begin{table}[t]
\caption{Recognition accuracy on Domain 5 after zero-shot domain adaptation.}\label{resfin}
\begin{center}
\resizebox{1.0\columnwidth}{!}{
\begin{tabular}{lccccccccc}
\hline 
Source ID  & 1 & 2 & 3 & 4 & 6 & 7 & 8 & 9 & Avg. \\
No DA & 66.12 & 67.33 & 66.69 & 71.73 & 69.74 & 69.82 & 68.68 & 62.93 & 67.88\\ 
ZSDA$\to$GFK & 67.90 & 69.03 & 69.03 & 74.43 & 71.80 & 72.66 & 71.09 & 66.41 & 70.29 \\ 
\hline 
\end{tabular} }
\end{center}
\scriptsize	 `Source ID' indicates \emph{which one} of the source domains is used to train the classifier.  `No DA' means applying the trained model directly to the target domain. `ZSDA$\to$GFK' is the proposed method: the subspace of the target domain is predicted by \emph{all} source domains, and the domain adaptation is conducted by GFK from each source domain in turn to the target domain. 
\end{table}

The results are summarised in Table~\ref{resfin}. These are generated by predicting the subspace $P_5$ of the unseen domain (domain\_5) based \emph{all} the remaining eight source domains and testing the classifier (with or without DA) trained from the labeled data of \emph{each} source domain in turn (i.e., single source domain to target). The source ID indicates which source domain is used to train the classifier and optionally apply DA. Note the distinction between source of unlabelled data for subspace learning, and source of labeled data for classifier learning. 

In every case, the proposed ZSDA method generates a useful subspace which allows domain adaptation to improve the performance.  It is as expected that using `closer' domains (measured by distance between $\mathbf{z}$'s) such as domain\_2, \_4, \_6, and \_8 to learn the classifier should lead to better performance compared to source domains that are `far-away' from the target. 
This is indeed the case: if we focus on the nearby domains\_$\{2,4,6,8\}$, then the average accuracy in Table~\ref{resfin} is 71.59\%. 
Given that the accuracy obtained within domain\_5 only (i.e. training and testing on domain\_5) is 72.94\%, and the average accuracy of those sources without DA is 69.37\%, this is a very encouraging result: \emph{we have reduced the cross-domain performance drop by over half without relying on any target domain data}.

\vspace{0.1cm}\noindent\textbf{Performance for each target domain:}\quad  
The previous analysis fixed domain\_5 as the target. We next extend this analysis and consider each domain in turn as the target. In each case we use the proposed algorithm to infer the subspace of the target domain given the eight sources, and we evaluate  the recognition accuracies with and without DA.
For conciseness, we just report the average accuracy over all possible label sources for each target (cf. the last column in Table~\ref{resfin}).
\begin{table}[t]
\caption{Average accuracy on every target domain after zero-shot domain adaptation.}
\label{resfin2}
\centering
\resizebox{1.0\columnwidth}{!}{
\begin{tabular}{lccccccccc}
\hline 
Target ID  & 1 & 2 & 3 & 4 & 5 & 6 & 7 & 8 & 9 \\
No DA & 69.10 & 69.15 & 68.31 & 68.18 & 67.88 & 63.23 & 59.14 & 56.86 & 46.41\\ 
ZSDA$\to$GFK & 72.06 & 72.07 & 71.24 & 71.07 & 70.29 & 65.94 & 61.37 & 59.54 & 50.76 \\
\hline 
\end{tabular} }
\end{table}
Table~\ref{resfin2} summarises the result and it demonstrates the effectiveness and robustness of the proposed method. It shows performance improvements on all choices of target domains, and on average, it boosts the accuracy by a factor of 4.75\%. 

\section{Conclusion}

We proposed the problem of continuously-parametrised zero-shot domain adaptation and developed a solution based on manifold-valued data regression. This allows us to predict the subspace to use at test-time and thus align a source classifier to a test-domain in advance of seeing any data. Preliminary results  demonstrate the value of our approach. This approach is highly promising for its potential impact on a variety of areas where it would be useful to be able to `calibrate' a recognition model on the fly based on metadata.

There are numerous areas for future work. A more thorough evaluation that also covers a wider range of applications is clearly needed. Our current kernel regression-based method is weak in domain extrapolation in contrast to interpolation. This is important for some kinds of subspace prediction, e.g., predicting future subspaces when one domain factor is time. Thus a generalisation for extrapolation is of interest. While the proposed method can and must use a set of available source domains to learn the subspace regressor, it can only exploit a single source domain's labels. A useful extension would therefore be to exploit multiple source domains worth of labels. Besides, it is interesting is to see if the predicted subspace can act as a regulariser so that it still helps when the target data are available but limited. Finally, a key assumption of this paper (in common with most other domain adaptation work) is that the domain descriptor is always observed and accurate. To relax this assumption and enable the model to deal with missing or noisy descriptors is also an interesting direction.

\bibliography{../../Library}
\end{document}